\documentclass[final,12pt]{elsarticle}
\let\today\relax
\makeatletter
\def\ps@pprintTitle{
    \let\@oddhead\@empty
    \let\@evenhead\@empty
    \def\@oddfoot{
\footnotesize\itshape
         {Submitted preprint} \hfill\today}\let\@evenfoot\@oddfoot
    }
\makeatother

\usepackage{amssymb}
\usepackage{amsmath}

\newcommand{\R}{\mathbb{R}}
\newcommand{\K}{\mathcal{K}}

\newcommand{\pkg}[1]{{\fontseries{m}\fontseries{b}\selectfont #1}}
\let\code=\texttt
\let\proglang=\textsf

\usepackage{booktabs}
\usepackage{siunitx}
\newcolumntype{d}{S[table-format=3.2]}

\usepackage{hyperref, lineno}

\begin{document}

\begin{frontmatter}

\title{Automated Model Selection for Generalized Linear Models}

\author[ict]{Benjamin Schwendinger\corref{cor1}} \ead{benjaminschwe@gmail.com}
\cortext[cor1]{Corresponding author.}
\author[aau]{Florian Schwendinger}
\author[cstat]{Laura Vana-G\"ur}
\affiliation[ict]{organization={Institute of Computer Technology, TU Wien},
            addressline={Gu{\ss}hausstra{\ss}e 27-29},
            city={Vienna},
            postcode={1040},
            country={Austria}}
\affiliation[aau]{organization={Department of Statistics, University of Klagenfurt},
            addressline={Universitätsstra{\ss}e 65-57},
            city={Klagenfurt},
            postcode={9020},
            country={Austria}}
\affiliation[cstat]{organization={Institute of Statistics and Mathematical Methods in Economics, TU Wien},
            addressline={Wiedner Hauptstra{\ss}e 7},
            city={Vienna},
            postcode={1040},
            country={Austria}}

\begin{abstract}
In this paper, we show how mixed-integer conic optimization can be used to combine feature subset selection with holistic generalized linear models to fully automate the model selection process.
Concretely, we directly optimize for the Akaike and Bayesian information criteria while imposing constraints designed to deal with multicollinearity in the feature selection task.
Specifically, we propose a novel pairwise correlation constraint that combines the sign coherence constraint with ideas from classical statistical models like Ridge regression and the OSCAR model.
\end{abstract}


\end{frontmatter}

\section{Introduction}
Model selection is an important but typically time-consuming task.
Several authors suggest using mixed-integer optimization to automate model selection.
The classical best subset selection (BSS) problem~\citep{subsetSelection:Miller:2002} identifies the best $k$ features out of $p$ possible features according to some goodness-of-fit (GOF) measure.
For linear regression, the BSS problem, in the least-squares sense is non-convex and known to be NP-hard~\citep{sparse:Natarajan:1995}.
In practice, the BSS problem is commonly approached using methods such as stepwise regression or exhaustive enumeration~\citep{subsetSelection:Miller:2002}.
\citet{bestSubset:Bertsimas:2016} highlight the significant advancements in mixed-integer optimization solvers since the 1970s when BSS problems were first investigated.
As a result, the BSS problem, which was once perceived as intractable, can now be solved by mixed-integer linear optimization solvers and used to find the solution of real-world linear regression problems.
In the case of logistic regression, the BSS problem becomes a mixed-integer non-linear optimization problem.
To address this, \citet{aloRegLogistic:Bertsimas:2017} explore the utilization of several general-purpose non-linear mixed-integer solvers.
Another approach to tackle the BSS problem in linear and logistic regression is presented by \citet{l0learn:Hazimeh:2020}, who employ a hybrid approach combining cyclic coordinate descent with a local combinatorial search.

An extension to the BSS problem is the integration of an information criterion (IC) such as the Akaike information criterion (AIC) or the Bayesian information criterion (BIC) into the optimization problem's objective. 
This enables the selection of not only the best features for a fixed $k$ but the overall optimal features according to the chosen goodness-of-fit measure.
This approach is commonly referred to in the optimization literature as feature subset selection (FSS).
Based on the division of subset selection algorithms into classes of filter, wrapper and embedded methods as done by \citet{wrapper:guyon:2003}, FSS classifies as an embedded method. 
FSS has been applied for various model classes including generalized linear models (GLMs).
\citet{fssMallows:Miyashiro:2015} apply FSS to linear regression using Mallows' $C_p$ statistic as a GOF measure.
By using intelligent branch-and-bound strategies, \citet{lmsubsets:Hofmann:2020} demonstrate the ability to solve the FSS problem for linear regression with 1,000s of observations and 100s of features within seconds. 
\citet{fssBinary:Sato:2016} propose a linear approximation approach to solve the FSS problem in logistic regression models.
They utilize a tangent line approximation of the logistic loss function, resulting in a mixed-integer linear problem. 
Likewise, \citet{bssPoisson:saishu:2021} suggest using a piecewise-linear approximation, solvable through mixed-integer linear programming, to address the FSS problem in Poisson models.
Many heuristic approaches for solving the FSS problem for GLMs are based on known metaheuristics.
These include simulated annealing~\citep{simann:debuse:1997}, genetic algorithms~(\citealt{genetic:yang:1998, pkg:glmulti:2020}), ant colony optimization~\citep{aco:chen:2010} and particle swarm optimization~\citep{pso:wang:2007}.

In this paper, we aim to provide a unified framework for performing automated model selection in GLMs by solving the FSS problem using mixed-integer conic programming and to extend existing approaches to FSS to effectively handle two common challenges in GLM optimization: separation and multicollinearity.
Furthermore, our paper is the first to propose utilizing conic optimization specifically for the FSS problem in Poisson regression models.

The proposed framework builds on the class of holistic generalized linear models (HGLMs) introduced by \citet{holiglm:Schwendinger:2022} in  package \pkg{holiglm} for \proglang{R}.
HGLMs extend linear, binomial and Poison GLMs by adding constraints designed to improve the model quality (e.g.,~restricting the number of variables entering the model, enforcing coherent signs of coefficients, etc.).
Instead of approximating the log-likelihood or its components, \pkg{holiglm} formulates the underlying optimization problems as conic optimization problems, providing a more reliable~\citep{logbin:schwendinger:2021} and accurate solution approach.
The parameters of these constrained models are obtained by using 
(mixed-integer) conic optimization.

While the framework in \cite{holiglm:Schwendinger:2022} allows for the specification of an upper bound on the number of variables to be selected in the model, this bound must still be predetermined prior to solving the underlying optimization problem.
In particular, we modify the framework of HGLMs for the purpose of FSS and directly integrate the AIC and BIC into the objective function rather than using the likelihood function as an objective. 
This allows us to obtain an exact solution to the FSS problem without resorting to piecewise-linear approximation methods while treating FSS for linear, binomial and Poison GLMs in a unified way.

As mentioned above, a primary focus of the paper is addressing the issue of separation and multicollinearity in the automated model selection process.
For tackling strong multicollinearity, we propose a novel pairwise correlation constraint that combines ideas from the sign coherence constraint \citep{integer:Carrizosa:2020} with a simultaneous restriction of equal-magnitude coefficients.
The idea of restricting coefficients to have the same magnitude can also be found in other statistical models, such as Ridge regression~\citep{ridge:hoerl:1970}, where the coefficients shrink equally or the OSCAR (Octagonal Shrinkage and Clustering Algorithm for Regression) model~\citep{oscar:bondell:2008}, where exact equality of clustered coefficients is required.
Separation on the other hand is characterized by extreme overlap or distinct separation in the data and can result in unbounded optimization problems that common solvers often fail to detect.
To ensure reliable results, it is therefore vital to verify the existence of solutions.
For the binomial family with logit, probit, log and complementary log-log link, this can be done with a linear program~\citep{separation:Konis:2007} which we employ to avoid wrongly reporting solutions for problems whose solution is actually not determinable.

Through extensive simulation studies, we demonstrate the feasibility and practicality of our approach.
The results clearly indicate that our proposed constraint outperforms existing methods, offering a more accurate and efficient model selection process for GLMs, effectively addressing the challenges posed by multicollinearity.

The remainder of this paper is structured as follows: 
Section~\ref{sec:model} introduces best subset selection, feature subset selection and holistic generalized linear models.
Section~\ref{sec:pitfalls} explores common pitfalls that arise when estimating certain GLMs, such as failure to converge or the nonexistence of a solution and possible solutions to mitigate these problems.
In Section~\ref{sec:fsshglm}, we present our proposed optimization problem for automated model selection in GLMs. The importance of dealing with multicollinearity when aiming for automated feature selection is illustrated through a simulation study in Section~\ref{sec:sim}.
Section~\ref{sec:conclusion} concludes the paper.

\section{Feature subset selection in GLMs}\label{sec:model}

In this section we set the stage for introducing the proposed modeling approach 
for performing FSS in GLMs. More specifically, we start with a brief introduction
to GLMs in Section~\ref{sec:methods:glm} and show how to formulate
the likelihood optimization problem such that it can be solved using
(mixed-integer) conic programming.
FSS is an extension to BSS, in the sense that the objective (i.e., log-likelihood) 
in the BSS problem is replaced by an information criterion. Before introducing
the FSS problem, we introduce the formulation of the BSS problem as a conic 
program in Section~\ref{sec:bss}. 
Section~\ref{sec:fss} introduces the information criteria we use to extend BSS to FSS.
\subsection{Generalized linear models and conic optimization}\label{sec:methods:glm}
In this work we focus on solving the FSS problem for GLMs with 
linear, binomial and Poisson families. Generally, GLMs as introduced by \citet{glm:nelder:1972}, are a classs of models with probability density functions that belong to the exponential dispersion model (EDM) family with probability density function:
\begin{equation}\label{eqn:pdf_edm}
f(y; \theta, \phi) = \exp\left(\frac{(y\theta - b(\theta))}{\phi} + c(y, \phi)\right).
\end{equation}
Here, $b(\cdot)$ and $c(\cdot)$ are well-defined functions that vary depending on the specific distribution.
In addition, in the presence of a (design) matrix of covariates $X$ with $p+1$ columns
(including a column of ones), a GLM has a linear predictor $\eta = X\beta$, and a link function $g$ that establishes the relationship between the linear combination of the $p+1$ covariates and the mean of response $y_i$: $g(E(y_i))=\eta_i$. 
Given $g$ and $b$, $\theta$ is then a function of $\eta$ and therefore of $\beta$. 
Given a sample of $n$ independent and identically distributed response observations
$y^\top = (y_1, \dots, y_n)$ and observed covariates, the estimation of the parameters is usually done by maximum likelihood,
and the maximum likelihood estimate (MLE) of $(\beta, \phi)$ are the values $(\beta^*, \phi^*)$ that maximize the (log)-likelihood function for the EDM family:
\begin{equation}  \label{eqn:loglik}
\log \mathcal{L}(\beta; y) = \sum_{i=1}^n \log f(y; \theta, \phi) = \sum_{i=1}^n \frac{y_i \theta_i(\beta) - b(\theta_i(\beta))}{\phi_i} + c(y_i, \phi_i).
\end{equation}

Conic optimization provides a framework for expressing the maximization of the 
log-likelihood function of various GLMs as convex optimization problems. 
A conic optimization problem is designed to model convex problems by
optimizing a linear objective function over the intersection of an affine hyperplane and a
nonempty closed convex cone.
The log-likelihood maximization can be reformulated as a conic problem.
The reason for this lies in the fact that the log-likelihood of common GLMs includes functions that can be represented by convex cones, which in turn can be solved by modern conic optimization solvers.
The estimation by means of the conic programming is in 
turn feasible given that efficient optimization solvers exist which can provide exact solutions. 
More specifically, the MLE for the linear regression model (Gaussian family with identity link) is the solution of a convex optimization problem which uses the second-order cone $\K^n_\text{soc} := \{(t,x) \in \R^n ~|~ x \in \R^{n-1}, t \in R, ||x||_2 \leq t\}$.
Since both logistic regression and Poisson regression involve exponential and logarithmic terms in their log-likelihoods, the primal exponential cone $\mathcal{K}_\text{expp} := \{(x,y,z) \in \R^3 ~|~ y > 0, ye^{\frac{x}{y}} \leq z \} \cup \{(x,0,z) \in \R^3 ~|~ x \leq 0, z \geq 0\}$ is utilized to represent them.

A comprehensive introduction to conic optimization can be found in~\citet{convex:boyd:2004}.
For the detailed explanation and derivation of conic formulations for various GLMs based on the family and link information, we refer to the appendix provided by \citet{holiglm:Schwendinger:2022}.

\subsection{Best subset selection}\label{sec:bss}
We introduce the classical best subset selection (BSS) problem before presenting its 
extension to FSS.
The classical best subset selection (BSS) problem is concerned with determining the best $k$ features from a set of $p$ potential features using some goodness-of-fit (GOF) metric.
However, the BSS problem is non-convex and NP-hard~\citep{sparse:Natarajan:1995}. 
Surrogate models are often used to overcome this computational burden, such as those that incorporate an $L_1$ penalty or a combination of $L_1$ and $L_2$ penalties~\citep{elastic:Zou:2005}.
Adding an $L_1$ penalty gives the least absolute shrinkage and selection operator (LASSO)~\citep{lasso:Tibshirani:1996}.
However, the performance of BSS and LASSO depends on the signal-to-noise ratio.
A comprehensive overview by \citet{bestSubset:Hastie:2020} highlights situations for the linear regression case where BSS outperforms LASSO and vice versa.
\citet{l0estimation:yang:2024} approximate the BSS problem by solving a sequence of weighted LASSO problems, with the weights determined progressively.

For generalized linear models where the log-likelihood is used as the goodness-of-fit measure, the BSS problem can be formulated as an optimization problem.
The goal is to minimize the negative log-likelihood (equivalent to maximizing the likelihood) with respect to the parameter vector $\beta$ while constraining the number of selected features to be $k$.
This results in the following optimization problem:
\begin{equation} \label{eqn:bssglm}
    \underset{\beta}{\text{minimize}} -\log \mathcal{L}(\beta; y)
    \quad \textrm{subject to} \quad
    \sum_{i=1}^p \mathbb{I}_{\{\beta_i \neq 0\}} \leq k.
\end{equation}
Here, $k \in {1, \dots, p}$ is a user-defined parameter that restricts the size of the subset.
We can formulate the whole problem as a mixed-integer convex optimization problem containing the constraint with the $\ell_0$ pseudo-norm, which counts the number of non-zero entries in $\beta$.
To do so, we introduce $p$ binary variables $z_i$ that indicate whether the covariate $\beta_i$ is selected for the model or not.
Note that $\beta_0$ denotes the intercept, which is always included in the model.
Now, the BSS problem becomes the following problem:
\begin{equation} \label{eqn:bssglmbigm}
\begin{array}{rl}
    \underset{\beta}{\text{minimize}} & - \log \mathcal{L}(\beta; y) \\
    \text{subject to} & -M z_i \leq \beta_i \leq M z_i, \quad i = 1, \dots, p, \\
                      & \sum\limits_{i=1}^p z_i \leq k, \\
                      & \beta \in \R^{p+1}, z \in \{0, 1\}^p.
\end{array}
\end{equation}
Here, $M \geq ||\hat{\beta}||_\infty$ is a constant that ensures that a coefficient $\beta_i$ is zero if the corresponding binary variable $z_i$ is zero.
In other words, $z_i$ indicates whether $\beta_i$ is included in the model. 
These types of constraints are often referred to as big-$M$ constraints.
It is well documented that problems containing big-$M$ constraints depend on a good choice of $M$.
If $M$ is too small, the convergence to the same optimum as the original problem is not guaranteed. 
If $M$ is too large, loss of accuracy and numerical instability may occur.
One should also be aware that choosing an arbitrarily large $M$ results in an unnecessarily large feasible region for the LP relaxation~\citep{bigm:Camm:1990}.
To address this challenge, \citet{bestSubset:Bertsimas:2016} proposed a data-driven method to determine lower and upper bounds for $\hat{\beta}_i$ in linear regression.
In the following, we extend this approach to convex GLMs.
Let UB be an upper bound on the MLE of Problem~\ref{eqn:bssglm}. 
Then one can find lower and upper bounds by solving the following convex optimization problems:
\begin{equation}
    \begin{array}{ll}
        u_i^+ := & \underset{\beta}{\text{maximize}} \quad \beta_i \\
        \text{s. t.} & -\log \mathcal{L}(\beta; y) \leq \text{UB},
    \end{array}
\end{equation}
\begin{equation}
    \begin{array}{ll}
        u_i^- := & \underset{\beta}{\text{minimize}} \quad \beta_i \\
        \text{s. t.} & -\log \mathcal{L}(\beta; y) \leq \text{UB}
    \end{array}
\end{equation}
where $u_i^+$ is an upper bound and $u_i^-$ is a lower bound to $\hat \beta_i$.
Now, let $M_i := \max\{|u_i^+|, |u_i^-|\}$ and one can choose the big-$M$ as $M = \underset{i}{\max} ~ M_i$.

This procedure involves solving several convex optimization problems and estimating an upper bound (UB) beforehand.
An alternative simpler approach, which we have also employed in our simulation studies, is to standardize the design matrix $X$ and choose an $M$ that works well for most settings.
We have found that a value of $M = 100$ works well for many data sets when using a standardized design matrix.
\subsection{Information criteria for feature subset selection}\label{sec:fss}
While BSS obtains the best subset of features for a fixed number of maximal active coefficients $k$ (where a coefficient is said to be active if it is non-zero), 
information criteria such as the Akaike Information Criterion~\citep{AIC:Akaike:1974} or the Bayesian Information Criterion~\citep{BIC:Schwarz:1978} are often used (typically in a second stage, see \citet{lmsubsets:Hofmann:2020})
to select the best model out of the candidate models with different number of active coefficients.
For linear regression, it is computationally advantageous not to optimize the AIC directly but to use Mallows' $C_p$ statistic (CP)~\citep{CP:Mallows:1964} instead. 
\citet{akaike:boisbunon:2014} show that for linear regression, the AIC and CP are equivalent in the sense that both reach their minimum objective value with the same set of active coefficients.

Instead of having a two stage procedure, in FSS we will replace the objective function in Equation~\ref{eqn:bssglmbigm} by an IC. Although many other ICs exist, we will focus on the AIC and BIC in this paper, as they are the most commonly used in practice.
For linear regression, however, we use the computationally advantageous CP to replace the AIC and a modified CP (CP2) to replace the BIC.
The proof of the equivalence of BIC and CP2 can be found in~\ref{app:cp2}.

The Akaike Information Criterion (AIC) is defined as follows:
\begin{equation}\label{eqn:aic}
    \text{AIC} = 2k - 2 \log(\mathcal{L})
\end{equation}
where $\mathcal{L}$ is the likelihood function and $k$ represents the number of active covariates.
Similarly, the Bayesian Information Criterion (BIC) is defined as:
\begin{equation}\label{eqn:bic}
    \text{BIC} = k \log(n) - 2 \log(\mathcal{L}).
\end{equation}
The Mallows' $C_p$ statistic (CP) for linear regression is defined by:
\begin{equation}\label{eqn:cp}
    \text{CP} = 2k + \frac{1}{\sigma^2} ||y - X \beta||_2^2 - n.
\end{equation}
Similarly, the modified CP, which is equivalent to the BIC, is defined as follows:
\begin{equation}\label{eqn:cp2}
    \text{CP2} = k \log(n) + \frac{1}{\sigma^2} ||y - X \beta||_2^2 - n.
\end{equation}
Here $\sigma^2$ is the residual variance.
The following observation, which is sometimes referred to as monotonicity of the GOF measure, applies to AIC, BIC, CP and CP2.
When we encounter two models with the same likelihood value, then the model with fewer selected coefficients is better.
We use the unbiased estimator of the variance $\hat{\sigma}^2 = \frac{||y - X \beta||^2_2}{n - p}$ from the full regression model.

\section{Issues in GLM optimization}\label{sec:pitfalls}
In this section, we investigate common data settings of GLMs that cause optimizers to recover wrong solutions.
Two main reasons are causing optimizers to recover an incorrect solution for GLMs.
Firstly, for GLMs with a binomial response, separation in the data can cause the maximum likelihood estimate to contain infinite components.
Secondly, cases where the data exhibits strong multicollinearity can lead to instability in the estimation and in inconsistent signs in the regression coeffients.

Before introducing our approach, we present in the following some proposed approaches
in the literature to address these issues.

\subsection{Separation}
\citet{mleExistence:Albert:1984} show that for logistic regression and probit models, the finiteness and uniqueness of the MLE are connected to the overlap
of the data.
They identify three different data settings, \textit{complete separation}, \textit{quasi-complete separation} and \textit{overlap} and show that for logistic regression overlap is necessary and sufficient for the finiteness of the MLE.
\citet{Safe:Konis:2009} translate these criteria into a linear problem, which can be checked to verify the existence of the MLE.

Although, in theory, the solvers should be able to detect the unboundedness of the problem, we found in our experiments that all the solvers we tried failed to detect the unboundedness for GLMs with a binomial response.
Therefore, it is important to check in the simulation that, indeed, a solution exists;
otherwise, we would mainly compare the default tolerance settings of the solvers.

\citet{Safe:Konis:2009} suggest using the following linear program (LP) to verify that the solution exists:
\begin{equation}
\label{eqn:lp_sep_logistic}
\begin{array}{rll}
  \underset{\beta}{\text{maximize}}~~  &
    \sum\limits_{i \in I} x^\top_i \beta - \sum\limits_{i \in J} x^\top_i \beta \\
  \text{subject to}~~  &
    x^\top_i \beta \leq 0 \quad \forall i \in J = \{i|y_i = 0\} \\
  & x^\top_i \beta \geq 0 \quad \forall i \in I =  \{i|y_i = 1\}. \\
\end{array}
\end{equation}
If the solution is a zero vector, this verifies that the data is overlapping and that the solution of the corresponding logistic regression model is finite and unique.
In case the solution of the LP is unbounded, the data is \mbox{(quasi-)}separated and the MLE does not exist.

In the preparation of this paper, we found more than two examples of peer-reviewed articles where the authors did not check their data for separation, which led them to report results based on unbounded optimization problems.
This likely occurred as a consequence of the solver incorrectly signaling convergence.

\subsection{Multicollinearity}\label{sec:mcollin}
In the presence of strong collinearity, the matrix $X^\top X$ in linear regression becomes ill-conditioned, leading to unstable estimates of the coefficients ($\hat{\beta}$).
This instability can result in inflated estimates and inconsistent signs of the coefficients.
Different approaches to solving this problem have been proposed in the optimization and statistics literature.
One approach, as suggested by \citet{algoReg:Bertsimas:2015}, is to limit collinearity by incorporating the \textit{at most one constraint}: 
\begin{equation}\label{eqn:at_most_one}
z_i + z_j \leq 1 \quad \forall (i, j) \in \mathcal{HC} = \{(i, j) : ~ \nu \leq |\rho_{ij}|\},
\end{equation}
where the binary $z$ variables are the ones introduced in Equation~\ref{eqn:bssglmbigm}, 
$\nu$ is a predefined constant and $\rho_{ij}$ denotes Pearson's empirical correlation coefficient between the $i$-th and $j$-th columns of the design matrix $X$. 
Hence, $\mathcal{HC}$ represents the set of highly correlated features.
This constraint limits the pairwise correlation by ensuring that, at most one of the variables among a pair with a correlation exceeding $\nu$ is selected in the regression model.

\citet{integer:Carrizosa:2020} relax this constraint to the \textit{sign coherence constraint}~(see Equation~\ref{eqn:sign_coherence})
\begin{equation}\label{eqn:sign_coherence}
-M(1-u_{ij}) \leq \beta_i, ~  \operatorname{sign}(\rho_{ij}) \beta_j \leq M u_{ij} \quad
    \forall (i, j) \in \mathcal{G} = \{(i, j) : ~ \tau \leq |\rho_{ij}| \} \\
\end{equation}
to force coefficients of covariates with large pairwise multicollinearity to have coherent signs.
Hence, highly positively correlated features must have coefficients with the same sign, while highly negatively correlated features must have coefficients with opposite signs. 
Again, $M$ is a sufficiently large enough constant and $u_{ij}$ is a binary variable introduced to enforce the sign coherence.
One can see that for $u_{ij}=1$ and positive $\rho_{ij}$ it holds that $0 \leq \beta_{i}, \beta_j \leq M$.
Similarly, for $u_{ij}=0$ and negative $\rho_{ij}$ we have that $-M \leq \beta_i \leq 0$ and $0 \leq \beta_j \leq M$.
Clearly, the \textit{sign coherence constraint} is less restrictive than the \textit{at most one constraint}.

On the other hand, a different but related approach in statistics is to assume that strongly correlated features should have similar estimates.
This assumption is utilized in models like Ridge regression~\citep{ridge:hoerl:1970} and the OSCAR model~\citep{oscar:bondell:2008}.
In Ridge regression, a $L_2$ penalty term is added to the objective function, encouraging similar estimates for strongly correlated features. 
The OSCAR model goes further by enforcing exactly the same coefficient for strongly correlated features, inducing a clustering behavior among the coefficients.

\section{Suggested model}\label{sec:fsshglm}
\subsection{FSS for the Poisson model}\label{sec:fsspoisson}
We present in this section the formulation of the FSS problem for the Poisson model
using the AIC, which, to the best of our knowledge, has not been proposed before in the literature. The corresponding AIC and BIC formulations for linear, logistic, and Poisson regression can be found in~\ref{app:fss}.

We formulate the following mixed-integer conic program for feature subset selection:
\begin{equation} \label{eqn:fss_poisson_aic}
    \begin{array}{rl}
        \underset{\beta, \delta}{\text{minimize}} & 2\left( \sum\limits_{j=1}^p z_j \right) - 2 \left( \sum\limits_{i=1}^n y_i x_i^\top \beta - \delta_i\right) \\
        \text{subject to} & (x_i^\top \beta, 1, \delta_i) \in \K_\text{expp}, \quad i = 1, \dots, n, \\
                          & -M z_i \leq \beta_i \leq M z_i, \quad i = 1, \dots, p, \\
                          & \beta \in \R^{p+1}, z \in \{0, 1\}^p, \delta \in \R^n.
    \end{array}
\end{equation}
This problem can be solved by off-the-shelf mixed-integer conic optimization solvers like \pkg{ECOS}~\citep{ecos:Domahidi:2013} or \pkg{MOSEK}~\citep{lib:mosek}.
It is worth noting that \citet{bssPoisson:saishu:2021} have highlighted the concave but non-linear nature of the log-likelihood function for Poisson regression and proposed a piecewise-linear approximation method. On the other hand, we are the first to suggest utilizing conic optimization to solve the feature subset selection problem specifically for Poisson regression.
Similar problem formulations can be established for all family-link combinations introduced in~\cite{holiglm:Schwendinger:2022}.

\subsection{Combined constraint for multicollinearity}\label{sec:combinedconstraint}
To further enhance this FSS model to handle multicollinearity, we propose the so-called
\textit{combined constraint} where we integrate the \textit{sign coherence constraint} and a \textit{equal magnitude constraint} as a unified criterion. Based on the idea of similar estimates for highly correlated features, we believe that an equal magnitude constraint would be beneficial in much the same way that the \textit{sign coherence constraint} extends the \textit{at most one constraint}.
Assuming that the design matrix $X$ has been standardized, we can define the \textit{equal magnitude constraint} as follows:
\begin{equation}
\beta_i = \operatorname{sign}(\rho_{ij}) \beta_j \quad \forall (i, j) \in \mathcal{HC} = \{(i, j) : ~ \nu \leq |\rho_{ij}| \}
\end{equation}
here $\rho_{ij}$ again refers to the Pearson's correlation coefficient, and $\nu$ is a predefined constant threshold. 
This constraint ensures that the coefficients of strongly correlated features have the same magnitude but possibly different signs based on the correlation direction.

This combination allows us to automatically control pairwise multicollinearity 
in addition to the feature subset selection process.
The \textit{sign coherence constraint} is applied to variables exhibiting a moderate to strong pairwise correlation, while the \textit{equal magnitude constraint} is exclusively imposed on strongly correlated pairs.
By employing the \textit{equal magnitude constraint} instead of the previous at most one constraint, users gain more insightful information about the underlying data structure, instead of making a near-random selection.

\subsection{Final model}
By incorporating FSS with our novel \textit{combined constraint}, we formulate the following optimization problem:
\begin{equation}
    \label{eqn:final_model}
\begin{array}{rl}
    \underset{\beta}{\text{minimize }} 
        & \quad \text{IC} \\[\smallskipamount]
    \textrm{subject to}
        & \quad -M z_i \leq \beta_i \leq M z_i, \quad i = 1, \dots, p, \\[\smallskipamount]
        & -M(1-u_{ij}) \leq \beta_i, ~  \operatorname{sign}(\rho_{ij}) \beta_j \leq M u_{ij} \quad \forall (i, j) \in \mathcal{G} \\[\smallskipamount]
        & \quad \beta_i = \operatorname{sign}(\rho_{ij}) \beta_j \quad \forall (i, j) \in \mathcal{HC} \\
        & \beta \in \R^{p+1}, z \in \{0, 1\}^p, u \in \{0,1\}^{|\mathcal{G}|}
\end{array}
\end{equation}
where $\mathcal{G} = \{(i, j) : ~ \tau \leq |\rho_{ij}| < \nu \}$ and $\mathcal{HC} = \{(i, j) : ~ \nu \leq |\rho_{ij}| \}$, with $0 \leq \tau < \nu \leq 1$.
Again, $M$ represents a sufficiently large positive constant.
This model simultaneously enforces coherent coefficient signs for moderately correlated features and equal coefficient magnitudes for highly correlated features.
The thresholds for medium and high correlations are defined by $\tau$ and $\nu$, respectively. 
Moreover, before estimating any binomial model, we employ the linear program in \cite{separation:Konis:2007} to identify the problems whose solution is actually not determinable.

It is worth noting that the objective function of our final model, as shown in Equation~\eqref{eqn:final_model}, incorporates an arbitrary information criterion.
Consequently, our model can accommodate multiple information criteria, such as AIC, BIC, CP, and CP2, which can be expressed using the respective formulas in Equation~(\ref{eqn:aic},~\ref{eqn:bic},~\ref{eqn:cp},~\ref{eqn:cp2}).
This not only allows for greater flexibility, but also opens to the door to more customized models.
The model can be further extended by using further holistic constraints and expert knowledge can be seamlessly integrated into the model.
Our entire approach should be viewed as an additional tool in the modern data scientist's tool belt.

\section{Simulation studies}\label{sec:sim}
In our simulation study, we present two key findings.
Firstly, our approach demonstrates the ability to recover the true predictors.
Specifically, we achieve selection accuracy comparable to that of exact methods, highlighting the quality of our solutions.
Secondly, our newly integrated constraint proves successful in estimating variables within a multicollinearity context.
This finding emphasizes the effectiveness of our approach in overcoming challenges posed by multicollinearity and recovering accurate estimates.

All computational experiments were conducted on a Dell XPS15 laptop with an Intel Core i7--8750H CPU @ 2.20GHzx12 processor and 32 GB of RAM.
We utilized three mixed-integer optimization solvers: \pkg{Gurobi} 9.1.2~\citep{gurobi}, \pkg{MOSEK} 10.0.34~\citep{lib:mosek}, and \pkg{ECOS} 2.0.5~\citep{ecos:Domahidi:2013}.
The \proglang{R} package \pkg{ROI} was employed for representing the optimization problems.

To obtain the exact reference solutions, we utilized the \proglang{R} packages \pkg{lmSubsets}~\citep{lmsubsets_pkg:Hofmann:2021} and \pkg{bestglm}~\citep{bestglm:McLeod:2020}.
The \pkg{bestglm} package can also solve linear regression problems, but we exclude it from the comparison as \pkg{lmSubsets} exhibits significantly faster performance.
There exist many more \proglang{R} packages for subset selection, such as \pkg{glmulti}~\citep{pkg:glmulti:2020}, \pkg{L0Learn}~\citep{l0learn:Hazimeh:2020} or \pkg{abess}~\citep{pkg:abess:2023}, but only the selected two ensure that the solutions are indeed globally optimal.
In order to ensure a fair comparison, all solvers were restricted to utilizing only a single core.
\subsection{Simulation without multicollinearity}
In this simulation, we compare the performance of the FSS formulations suggested in Equation~(\ref{eqn:fss_poisson_aic},~\ref{eqn:fss_gaussian_aic},~\ref{eqn:fss_binomial_aic},~\ref{eqn:fss_gaussian_bic},~\ref{eqn:fss_binomial_bic},~\ref{eqn:fss_poisson_bic}) with special purpose solvers for FSS and BSS.

We use true positives ($TP$) and true negatives ($TN$) to calculate the selection accuracy. 
The true positives are the number of features $j$ for which both the estimated coefficient ($\hat{\beta_j}$) and the true coefficient ($\beta_j^\text{true}$) are non-zero:
\begin{equation}
TP(\beta) = |{j : \hat{\beta_j} \neq 0, \beta_j^\text{true} \neq 0}|.
\end{equation}
Analogously, the true negatives represent the number of features $j$ for which both the estimated coefficient ($\hat{\beta_j}$) and the true coefficient ($\beta_j^\text{true}$) are zero:
\begin{equation}
TN(\beta) = |{j : \hat{\beta_j} = 0, \beta_j^\text{true} = 0}|.
\end{equation}
Once $TP$ and $TN$ are calculated, the selection accuracy $(A)$ is determined as the sum of $TP$ and $TN$ divided by the total number of potential features ($p$):
\begin{equation}
A(\beta) = \frac{TP + TN}{p}.
\end{equation}
The runtime provides an indication of the computational efficiency of the solvers.
At the same time, the selection accuracy measures how well the solvers are able to correctly identify the relevant and irrelevant features.
For linear regression, we employ the specialized solver \pkg{lmSelect} from the \pkg{lmSubsets} package, which utilizes a branch-and-bound strategy tailored for this problem.
For logistic and Poisson regression, we use the dedicated solver \pkg{bestglm}.
Unlike \pkg{lmSubsets}, \pkg{bestglm} employs complete enumeration and is only suitable for regression problems with a moderate number of features.
To avoid excessively long runtimes, \pkg{bestglm} restricts the maximum number of features (for non-Gaussian families) to 15.

Following the setting of \citet{lmsubsets:Hofmann:2020} the simulation study adopts the following design: 
the design matrix $X$ is generated from a multivariate normal distribution, with $x_i \sim \mathcal{N}(0, \Sigma)$ for $i = 1, \dots, n$ with a mean of zero and the covariance matrix $\Sigma$ is the identity matrix $I_p$. 
Each scenario consists of 5 different runs, with $n = 1000$ observations.
The number of features $p$ varies, and for each scenario, the first $\lceil \frac{p}{2} \rceil$ coefficients of $\beta$ are set to 1, while the remaining coefficients are set to 0.

In the linear regression setting, we generate 175 datasets.
The number of features $p$ varies among $\{20, 25, 30, 35, 40, 45, 50\}$, and different standard deviations $\sigma \in \{0.05, 0.10, 0.50, 1.00, 5.00\}$ are employed.
The response variable $y_i$ is generated according to $y_i = x_i^\top \beta + \epsilon_i$, where $\epsilon_i$ follows a normal distribution with mean 0 and variance $\sigma^2$.
To avoid long runtimes in the brute force setting, we limit the allowed time for each run to 4200 seconds.

In the logistic and Poisson regression settings, we vary the number of features within $p \in \{5, 10, 15, 20, 25, 30, 35, 40\}$.
The inverse link function is used to calculate $\mu_i = g^{-1}(\eta_i)$, where $\eta_i = x_i^\top \beta$.
Depending on the model, the response $y_i$ is sampled from either $y_i \sim \textrm{Binomial}(1, p = \mu_i)$ or $y_i \sim \textrm{Poisson}(\lambda = \mu_i)$, based on the underlying distribution assumption.

Tables~\ref{tab:gauss_fss_time}--\ref{tab:poisson_fss_sel} summarize the results of the computation time and selection accuracy for the different GLMs.

For linear regression, Table~\ref{tab:gauss_fss_sel} shows that all tested methods achieve a high selection accuracy.
However, in some cases, $\code{lmSelect}$ finds models with a slightly lower AIC or BIC, including more features.
Upon inspecting the data, it is discovered that the true coefficients of the additionally selected features are all 0.
This discrepancy can be attributed to the fact that the residual variance $\sigma^2$ is only estimated, and as a result, the minima of AIC and Mallows' $C_p$ may not coincide perfectly.
As for the computation time, Table~\ref{tab:gauss_fss_time} indicates that our FSS model is only slightly slower than $\code{lmSelect}$. 
Compared to the enumerating BSS approach of trying all different values for the sparsity parameter $k$, the FSS approach is faster and scales better with the number of features.

Regarding logistic regression, FSS can recover the actual coefficients with high accuracy, as can be seen in Table~\ref{tab:binomial_fss_sel}. 
In terms of scalability, our FSS approach scales better than the complete enumeration of $\code{bestglm}$. See Table~\ref{tab:binomial_fss_time} for the full timings.

Our findings for Poisson regression are similar to those for logistic regression. 
The computation time and selection quality results are summarized in Table~\ref{tab:poisson_fss_time} and Table~\ref{tab:poisson_fss_sel}.
Here we achieve excellent selection accuracies and FSS scales better than complete enumeration.
For Poisson regression, we found that \pkg{ECOS} encounters numerical problems for the simulation setting with more than 20 features (see also Table~\ref{tab:poisson_fss_sel}).

\subsection{Simulation with multicollinearity}
In this section, we compare the performance of FSS with and without pairwise correlation constraints. 
We use the mean squared error (MSE) as a performance measure of how well a method performs at recovering the estimates.
We consider three types of pairwise correlation constraints: the \textit{at most one constraint}, the \textit{sign coherence constraint} and the \textit{combined constraint}.
Moreover, to establish a baseline, we also benchmark a model selected based on the information criterion alone without additional constraints which we call \textit{no constraint} model.
Following the recommendation of \citet{algoReg:Bertsimas:2015} we set the threshold for the \textit{at most one constraint} to $\nu = 0.7$.
If the correlation between two features exceeds $0.7$, at most, one of the features can be selected in the feature subset.
The \textit{sign coherence constraint} evaluates two thresholds, $\tau = 0.5$ and $\tau = 0.7$.
This constraint ensures that if two features have a correlation magnitude above the threshold, they must have a coherent sign in the feature subset.
Thus, highly positively correlated features are forced to have the same sign, while highly negatively correlated features must have opposite signs.
The \textit{combined constraint} combines the ideas of the \textit{equal magnitude constraint} and the \textit{sign coherence constraint} by enforcing the equal magnitude for highly correlated feature pairs and ensuring sign coherence for moderately correlated feature pairs.
To distinguish between moderately and highly correlated feature pairs, we used the thresholds $\tau = 0.5$ and $\nu = 0.7$.

It is important to note that the simulation study design will impact the study's outcome, especially when comparing pairwise correlation constraints.
By using the simulation setting proposed by \citet{sim:McDonald:1975}, we aim to provide a common ground for comparing the constraints. 
This simulation setting is widely used in statistics to simulate collinearity among features.

In this setting, the design matrix $X$ is constructed as follows:
\begin{equation}
    x_{ij} = \sqrt{(1 - \alpha^2)} z_{ij} + \alpha_{i(p+1)}  \text{, \quad where } i = 1, \dots, n \text{ and } j = 1, \dots, p.
\end{equation}
Here, $n = 100$ represents the number of observations, $p = 3$ denotes the number of features, and $z_{ij}$ is drawn from a normal distribution with mean 0 and standard deviation $\sigma$. The design matrix $X$ is standardized, resulting in $X^\top X$ being in correlation form.
The parameter $\alpha$ takes values from the set $\{0.7, 0.8, 0.9, 0.95, 0.99\}$.
In this particular setup, $\rho_{ij} = \alpha^2$ gives the correlation between any two features.
Consequently, the pairwise correlations are $\rho_{ij} = \{0.49, 0.64, 0.81, 0.90, 0.98\}$.
The $\beta$ coefficients are generated by computing the eigenvectors of $X^\top X$ and selecting the eigenvector associated with the largest eigenvalue.
\begin{equation}\label{eqn:calc_eta}
    \eta_i = \beta_0 + \beta_1 x_{1i} + \beta_2 x_{2i} + \cdots + \beta_p x_{pi} = x_i^\top \beta
\end{equation}
Subsequently, Equation~\eqref{eqn:calc_eta} is employed to calculate $\eta_i$ where $\beta_0$ is set to zero and $\mu_i$ is given by $\mu_i = g^{-1}(\eta_i)$. 
Finally, the response $y_i$ is sampled either from
$y_i \sim \text{Normal}(\mu_i, \sigma)$ or
$y_i \sim \text{Binomial}(1, \mu_i)$ or
$y_i \sim \text{Poisson}(\mu_i)$.

This simulation setup allows for the generation of data with controlled correlations among features, enabling the evaluation and comparison of the different pairwise correlation constraints in the context of feature subset selection.
Table~\ref{tab:sim2_binomial}, Table~\ref{tab:sim2_poisson} and Table~\ref{tab:sim2_gaussian} summarize the results for the different constraints and correlation settings for 1,000 simulations and 100 observations.

In the linear regression setting, the AIC and BIC exhibited almost identical results, leading us to report only the AIC outcomes. 
Analysis of Table~\ref{tab:sim2_gaussian} suggests that when the standard deviation $\sigma$ is very small (specifically, $\sigma = 0.01$) and the $\alpha$ values range from small to medium (0.7, 0.8, and 0.9), both the \textit{no constraint} and the \textit{sign coherence constraint} slightly outperform the \textit{combined constraint}.
The \textit{at most one constraint} performs best for small correlations.
This observation can be attributed to the decreased likelihood of the constraint becoming active at lower $\alpha$ values, rendering its performance similar to the one of the \textit{no constraint}. 
Overall, the simulation results indicate that the \textit{combined constraint} performs exceptionally well in this simulation setting, effectively harnessing the advantages of the \textit{sign coherence constraint} and the \textit{equal magnitude constraint}.

Table~\ref{tab:sim2_binomial} presents the results for the logistic regression setting, where the generated $\mu_i$ values implicitly determine the standard deviation.
In this scenario, the \textit{combined constraint} outperformed all other constraints.
While the MSE values for all constraints were similar under low correlation conditions, the \textit{combined constraint} exhibited significantly lower MSE values compared to the other constraints under strong correlation.

Similarly, Table~\ref{tab:sim2_poisson} showcases the results for the Poisson regression setting.
In this case, all constraints performed similarly under low correlation conditions.
However, as the correlation increased, the \textit{combined constraint} emerged as advantageous in this simulation setting.
Furthermore, the results indicate that the MSE values for the \textit{no constraint} and the \textit{sign coherence constraint} were almost identical, suggesting that inconsistent signs were rare in this simulation setup.

\section{Conclusion}\label{sec:conclusion}
This paper proposes an automated model selection approach by combining FSS with a correlation constraint for specific types of generalized linear models (GLMs), including linear, logistic, and Poisson regression.
Our method utilizes conic optimization and information criteria such as AIC or BIC for feature subset selection (FSS).
Moreover, it also enables the integration of additional constraints such as limiting pairwise correlation.
We have shown that our approach achieves high selection accuracy and improves computational efficiency as the dimensionality of the problem increases, outperforming naive enumeration methods.
The key contribution of our work is the development of a new mixed-integer conic programming (MICP) problem for model selection under multicollinearity.
Our experiments have demonstrated that our proposed combined constraint surpasses the existing collinearity constraints in the literature regarding performance.
Future research directions could explore incorporating additional regularization terms or leveraging linear approximations to speed up computation times.

\section*{Acknowledgements}\label{sec:acknowledgements}
This work was supported by the Austrian Science Fund (FWF) under grant number ZK 35.

\appendix

\section{GLMs}\label{app:glm}
The PDF of the Normal distribution can be expressed as:
\begin{equation}
f(y; \mu, \sigma^2) = \frac{1}{\sqrt{2 \pi \sigma^2} \exp\left(\frac{-(y-\mu)^2}{2 \sigma^2}\right)} = \exp\left(\frac{(y \mu - \mu^2/2)}{\sigma^2} + \left(-\frac{y^2}{2 \sigma^2} - \frac{1}{2} \log(2 \pi \sigma^2)\right)\right)
\end{equation}
where one can see that it is indeed part of the EDM family with $\theta = \mu$, $\phi = \sigma^2$, $b(\theta) = \frac{\theta^2}{2}$
and $c(y, \phi) = -\frac{y^2}{2 \phi} - \frac{1}{2} \log(2 \pi \phi)$.

\noindent
The PDF of the binomial distribution can be written as:
\begin{equation}
f(y; p) = {n \choose y} p^y (1-p)^{n-y} = \exp\left(\log{n \choose y} + y \log(\frac{p}{1-p})+ n \log(1-p) \right).
\end{equation}
Together with the link function $\theta = \textrm{logit}(p) = \log\left(\frac{p}{1-p}\right)$ we get
\begin{equation}
f(y; p) = \exp\left(y \theta + n \log\left(\frac{1}{1 + \exp(\theta)}\right) + \log{n \choose y}\right).
\end{equation}
Here we can see that the binomial distribution is part of the EDM family with
$\theta = \log\left(\frac{p}{1-p}\right)$, $\phi = 1$, $b(\theta) = -n \log\left(\frac{1}{1 + \exp(\theta)}\right)$ and $c(y, \phi) = \log{n \choose y}$.

The PDF of the Poisson distribution can be expressed by:
\begin{equation}
f(y; \mu) = \frac{\mu^y \exp(-\mu)}{y!} = \exp(y \log(\mu) - \mu - \log(y!)).
\end{equation}
Now using the link function $\theta = \log(\mu)$ which is equal to $\exp(\theta) = \mu$, we get
\begin{equation}
f(y; \mu) = \exp(y \theta - \exp(\theta) - \log(y!)).
\end{equation}
We can see that the Poisson distribution is also part of the EDM family
with $\theta = \log(\mu)$, $\phi = 1$, $b(\theta) = \exp(\theta)$ and $c(y, \phi) = - \log(y!)$.

\section{Mallows $C_p$ for BIC}\label{app:cp2}
The log-likelihood for linear regression is given by:
\begin{equation}
\log \mathcal{L}(\beta; y) = - \frac{1}{2 \sigma^2} (y_i - x_i^\top \beta)^2 - \frac{1}{2} \log(2 \pi \sigma^2).
\end{equation}
Given the definition of BIC (in Equation~\eqref{eqn:bic}) and CP2 (in Equation~\eqref{eqn:cp2}) we have
\begin{equation}
\begin{array}{rl}
\textrm{BIC} = & k \log(n) - 2 \log(\mathcal{L}) \\
           = & k \log(n) + \sum_{i=1}^n \frac{1}{\sigma^2} (y_i - x_i^\top \beta)^2 + \log(2 \pi \sigma^2) \\
           = & k \log(n) + n \log(2 \pi \sigma^2) + \frac{||y - X \beta ||_2^2}{\sigma^2} \\
           = & \textrm{CP2} + n(\log(2 \pi \sigma^2) + 1)
\end{array}
\end{equation}
This implies that they reach their respective minima at the same coefficient values.

\section{Mixed-integer conic optimization problems for FSS}\label{app:fss}
\subsection*{FSS for AIC}
\subsubsection*{Gaussian}
\begin{equation}
    \begin{array}{rl} \label{eqn:fss_gaussian_aic}
    \underset{\beta, \zeta}{\text{minimize}} ~ & 2 \left(\sum\limits_{j=1}^p z_j \right) + \frac{1}{\sigma^2} \zeta \\
    \text{subject to} ~ 
    & ( \zeta + 1, \zeta - 1, 2( y_1 -   x_1^\top \beta ), \ldots, 2( y_n -   x_n^\top \beta )) \in \K_\text{soc}^{n+2} \\
    & -M z_i \leq \beta_i \leq M z_i, \quad i = 1, \dots, p, \\
    & \beta \in \R^{p+1}, z \in \{0, 1\}^p, \zeta \in \R.
    \end{array}
\end{equation}

\subsubsection*{Binomial}
\begin{equation}
    \begin{array}{rl} \label{eqn:fss_binomial_aic}
    \underset{\beta,\delta,\gamma}{\text{minimize}} & 2 \left( \sum\limits_{j=1}^p z_j \right) - 2 \left( \sum\limits_{i=1}^n y_i x_{i}^\top \beta - \delta_i \right)\\
    \text{subject to } 
        & (\delta_i, 1, 1+\gamma_i)\in \mathcal{K}_\text{expp}, \quad i = 1, \ldots, n \\
        & ( x_{i}^\top \beta, 1, \gamma_i)\in \mathcal{K}_\text{expp}, \quad i = 1, \ldots, n \\
        & -M z_i \leq \beta_i \leq M z_i, \quad i = 1, \dots, p, \\
        & \beta \in \R^{p+1}, z \in \{0, 1\}^p, \delta \in \R^{n},  \gamma \in \R^{n}.
    \end{array}
\end{equation}

\subsubsection*{Poisson}
The optimization problem is already given in Equation~\eqref{eqn:fss_poisson_aic}.

\subsection*{FSS for BIC}

\subsubsection*{Gaussian}
\begin{equation}
\begin{array}{rl} \label{eqn:fss_gaussian_bic}
    \underset{\beta, \zeta}{\text{minimize}} ~ & \log(n) \left(\sum\limits_{j=1}^p z_j \right) + \frac{1}{\sigma^2} \zeta \\
    \text{subject to} ~ 
    & ( \zeta + 1, \zeta - 1, 2( y_1 -   x_1^\top \beta ), \ldots, 2( y_n -   x_n^\top \beta )) \in \K_\text{soc}^{n+2} \\
    & -M z_i \leq \beta_i \leq M z_i, \quad i = 1, \dots, p, \\
    & \beta \in \R^{p+1}, z \in \{0, 1\}^p, \zeta \in \R.
    \end{array}
\end{equation}

\subsubsection*{Binomial}
\begin{equation}
    \begin{array}{rl} \label{eqn:fss_binomial_bic}
    \underset{\beta,\delta,\gamma}{\text{minimize}} & \log(n) \left( \sum\limits_{j=1}^p z_j \right) - 2 \left( \sum\limits_{i=1}^n y_i x_{i}^\top \beta - \delta_i \right)\\
    \text{subject to } 
        & (\delta_i, 1, 1+\gamma_i)\in \mathcal{K}_\text{expp}, \quad i = 1, \ldots, n \\
        & ( x_{i}^\top \beta, 1, \gamma_i)\in \mathcal{K}_\text{expp}, \quad i = 1, \ldots, n \\
        & -M z_i \leq \beta_i \leq M z_i, \quad i = 1, \dots, p, \\
        & \beta \in \R^{p+1}, z \in \{0, 1\}^p, \delta \in \R^{n},  \gamma \in \R^{n}.
    \end{array}
\end{equation}

\subsubsection*{Poisson}
\begin{equation} \label{eqn:fss_poisson_bic}
    \begin{array}{rl}
        \underset{\beta, \delta}{\text{minimize}} & \log(n) \left( \sum\limits_{j=1}^p z_j \right) - 2 \left( \sum\limits_{i=1}^n y_i x_i^\top \beta - \delta_i \right) \\
        \text{subject to} & (x_i^\top \beta, 1, \delta_i) \in \K_\text{expp}, \quad i = 1, \dots, n, \\
                          & -M z_i \leq \beta_i \leq M z_i, \quad i = 1, \dots, p, \\
                          & \beta \in \R^{p+1}, z \in \{0, 1\}^p, \delta \in \R^n.
    \end{array}
\end{equation}
\clearpage

\section*{Result Tables}
\begin{table}[htb!]

\caption{\label{tab:gauss_fss_time}Simulation without multicollinearity: Comparison of proposed method, HLM with enumeration and special purpose solver for linear regression; average execution times in seconds.}
\centering
\resizebox{\ifdim\width>\linewidth\linewidth\else\width\fi}{!}{
\begin{tabular}[t]{r>{}r|rr>{}r|rrr}
\toprule
\multicolumn{8}{c}{Linear model training time} \\
\cmidrule(l{3pt}r{3pt}){1-8}
\multicolumn{2}{c}{ } & \multicolumn{3}{c}{AIC} & \multicolumn{3}{c}{BIC} \\
\cmidrule(l{3pt}r{3pt}){3-5} \cmidrule(l{3pt}r{3pt}){6-8}
\multicolumn{2}{c}{ } & \multicolumn{1}{c}{Proposed method} & \multicolumn{1}{c}{ } & \multicolumn{1}{c}{HLM with brute force} & \multicolumn{1}{c}{Proposed method} & \multicolumn{1}{c}{ } & \multicolumn{1}{c}{HLM with brute force} \\
\multicolumn{1}{c}{$\sigma$} & \multicolumn{1}{c}{p} & \multicolumn{1}{c}{(Gurobi)} & \multicolumn{1}{c}{lmSelect} & \multicolumn{1}{c}{(Gurobi)} & \multicolumn{1}{c}{(Gurobi)} & \multicolumn{1}{c}{lmSelect} & \multicolumn{1}{c}{(Gurobi)} \\
\cmidrule(l{3pt}r{3pt}){1-1} \cmidrule(l{3pt}r{3pt}){2-2} \cmidrule(l{3pt}r{3pt}){3-3} \cmidrule(l{3pt}r{3pt}){4-4} \cmidrule(l{3pt}r{3pt}){5-5} \cmidrule(l{3pt}r{3pt}){6-6} \cmidrule(l{3pt}r{3pt}){7-7} \cmidrule(l{3pt}r{3pt}){8-8}
0.05 & 20 & 0.049 & 0.009 & 1.140 & 0.047 & 0.002 & 1.140\\
 & 25 & 0.063 & 0.003 & 2.989 & 0.062 & 0.003 & 2.989\\
 & 30 & 0.085 & 0.003 & 9.551 & 0.069 & 0.003 & 9.551\\
 & 35 & 0.196 & 0.004 & 58.876 & 0.086 & 0.004 & 58.876\\
 & 40 & 0.179 & 0.004 & 224.310 & 0.094 & 0.004 & 224.310\\
 & 45 & 0.214 & 0.007 & 1384.208 & 0.123 & 0.005 & 1384.208\\
 & 50 & 0.381 & 0.007 & -- & 0.149 & 0.006 & --\\
\hline\addlinespace
0.1 & 20 & 0.059 & 0.003 & 1.252 & 0.055 & 0.003 & 1.252\\
 & 25 & 0.065 & 0.003 & 3.347 & 0.060 & 0.003 & 3.347\\
 & 30 & 0.084 & 0.004 & 10.526 & 0.074 & 0.003 & 10.526\\
 & 35 & 0.169 & 0.005 & 51.952 & 0.111 & 0.005 & 51.952\\
 & 40 & 0.215 & 0.005 & 198.132 & 0.106 & 0.005 & 198.132\\
 & 45 & 0.229 & 0.006 & 1441.896 & 0.120 & 0.006 & 1441.896\\
 & 50 & 1.012 & 0.007 & -- & 0.185 & 0.008 & --\\
\hline\addlinespace
0.5 & 20 & 0.070 & 0.003 & 1.187 & 0.059 & 0.003 & 1.187\\
 & 25 & 0.073 & 0.004 & 2.900 & 0.076 & 0.004 & 2.900\\
 & 30 & 0.102 & 0.004 & 10.476 & 0.084 & 0.004 & 10.476\\
 & 35 & 0.141 & 0.005 & 54.417 & 0.113 & 0.005 & 54.417\\
 & 40 & 0.194 & 0.006 & 206.109 & 0.122 & 0.006 & 206.109\\
 & 45 & 0.353 & 0.007 & 1411.889 & 0.164 & 0.007 & 1411.889\\
 & 50 & 0.397 & 0.007 & -- & 0.146 & 0.007 & --\\
\hline\addlinespace
1 & 20 & 0.057 & 0.003 & 1.239 & 0.049 & 0.002 & 1.239\\
 & 25 & 0.063 & 0.004 & 3.074 & 0.076 & 0.003 & 3.074\\
 & 30 & 0.085 & 0.005 & 11.753 & 0.071 & 0.004 & 11.753\\
 & 35 & 0.124 & 0.005 & 65.864 & 0.091 & 0.004 & 65.864\\
 & 40 & 0.283 & 0.005 & 202.285 & 0.117 & 0.005 & 202.285\\
 & 45 & 0.275 & 0.007 & 1490.630 & 0.126 & 0.006 & 1490.630\\
 & 50 & 0.400 & 0.007 & -- & 0.143 & 0.008 & --\\
\hline\addlinespace
5 & 20 & 0.086 & 0.003 & 1.314 & 0.075 & 0.003 & 1.314\\
 & 25 & 0.095 & 0.004 & 3.085 & 0.093 & 0.004 & 3.085\\
 & 30 & 0.126 & 0.004 & 11.729 & 0.133 & 0.005 & 11.729\\
 & 35 & 0.154 & 0.004 & 45.243 & 0.164 & 0.006 & 45.243\\
 & 40 & 0.433 & 0.006 & 155.281 & 0.254 & 0.010 & 155.281\\
 & 45 & 0.309 & 0.007 & 1231.948 & 0.606 & 0.020 & 1231.948\\
 & 50 & 0.388 & 0.008 & -- & 1.151 & 0.035 & --\\
\bottomrule
\end{tabular}}
\end{table}
 \begin{table}[htb!]

\caption{\label{tab:gauss_fss_sel}Simulation without multicollinearity: Comparison of proposed method, HLM with enumeration and special purpose solver for linear regression; average (sd) selection accuracy.}
\centering
\resizebox{\ifdim\width>\linewidth\linewidth\else\width\fi}{!}{
\begin{tabular}[t]{r>{}r|rr>{}r|rrr}
\toprule
\multicolumn{8}{c}{Linear model support recovery accuracy} \\
\cmidrule(l{3pt}r{3pt}){1-8}
\multicolumn{2}{c}{ } & \multicolumn{3}{c}{AIC} & \multicolumn{3}{c}{BIC} \\
\cmidrule(l{3pt}r{3pt}){3-5} \cmidrule(l{3pt}r{3pt}){6-8}
\multicolumn{2}{c}{ } & \multicolumn{1}{c}{Proposed method} & \multicolumn{1}{c}{ } & \multicolumn{1}{c}{HLM with brute force} & \multicolumn{1}{c}{Proposed method} & \multicolumn{1}{c}{ } & \multicolumn{1}{c}{HLM with brute force} \\
\multicolumn{1}{c}{$\sigma$} & \multicolumn{1}{c}{p} & \multicolumn{1}{c}{(Gurobi)} & \multicolumn{1}{c}{lmSelect} & \multicolumn{1}{c}{(Gurobi)} & \multicolumn{1}{c}{(Gurobi)} & \multicolumn{1}{c}{lmSelect} & \multicolumn{1}{c}{(Gurobi)} \\
\cmidrule(l{3pt}r{3pt}){1-1} \cmidrule(l{3pt}r{3pt}){2-2} \cmidrule(l{3pt}r{3pt}){3-3} \cmidrule(l{3pt}r{3pt}){4-4} \cmidrule(l{3pt}r{3pt}){5-5} \cmidrule(l{3pt}r{3pt}){6-6} \cmidrule(l{3pt}r{3pt}){7-7} \cmidrule(l{3pt}r{3pt}){8-8}
0.05 & 20 & 0.91 (0.02) & 0.91 (0.02) & 0.91 (0.02) & 1.00 (0.00) & 1.00 (0.00) & 1.00 (0.00)\\
 & 25 & 0.90 (0.04) & 0.90 (0.04) & 0.90 (0.04) & 1.00 (0.00) & 1.00 (0.00) & 1.00 (0.00)\\
 & 30 & 0.89 (0.05) & 0.89 (0.05) & 0.89 (0.05) & 0.99 (0.03) & 0.99 (0.03) & 0.99 (0.03)\\
 & 35 & 0.90 (0.09) & 0.90 (0.09) & 0.90 (0.09) & 0.99 (0.01) & 0.99 (0.01) & 0.99 (0.01)\\
 & 40 & 0.92 (0.03) & 0.92 (0.04) & 0.92 (0.04) & 1.00 (0.00) & 1.00 (0.00) & 1.00 (0.00)\\
 & 45 & 0.90 (0.03) & 0.90 (0.03) & 0.90 (0.03) & 1.00 (0.01) & 1.00 (0.01) & 1.00 (0.01)\\
 & 50 & 0.94 (0.03) & 0.94 (0.03) & -- & 1.00 (0.00) & 1.00 (0.00) & --\\
\hline\addlinespace
0.1 & 20 & 0.90 (0.05) & 0.89 (0.07) & 0.90 (0.05) & 1.00 (0.00) & 1.00 (0.00) & 1.00 (0.00)\\
 & 25 & 0.93 (0.05) & 0.93 (0.05) & 0.93 (0.05) & 1.00 (0.00) & 1.00 (0.00) & 1.00 (0.00)\\
 & 30 & 0.93 (0.04) & 0.93 (0.04) & 0.93 (0.04) & 0.99 (0.01) & 0.99 (0.01) & 0.99 (0.01)\\
 & 35 & 0.91 (0.04) & 0.90 (0.05) & 0.91 (0.05) & 0.99 (0.01) & 0.99 (0.03) & 0.99 (0.03)\\
 & 40 & 0.93 (0.04) & 0.93 (0.04) & 0.93 (0.04) & 0.99 (0.01) & 0.99 (0.01) & 0.99 (0.01)\\
 & 45 & 0.91 (0.05) & 0.91 (0.05) & 0.91 (0.05) & 1.00 (0.01) & 1.00 (0.01) & 1.00 (0.01)\\
 & 50 & 0.90 (0.03) & 0.90 (0.03) & -- & 0.99 (0.01) & 0.99 (0.01) & --\\
\hline\addlinespace
0.5 & 20 & 0.91 (0.11) & 0.91 (0.11) & 0.91 (0.11) & 1.00 (0.00) & 1.00 (0.00) & 1.00 (0.00)\\
 & 25 & 0.91 (0.08) & 0.91 (0.08) & 0.91 (0.08) & 1.00 (0.00) & 1.00 (0.00) & 1.00 (0.00)\\
 & 30 & 0.96 (0.04) & 0.96 (0.04) & 0.96 (0.04) & 1.00 (0.00) & 1.00 (0.00) & 1.00 (0.00)\\
 & 35 & 0.93 (0.06) & 0.93 (0.06) & 0.93 (0.06) & 0.99 (0.01) & 0.99 (0.01) & 0.99 (0.01)\\
 & 40 & 0.92 (0.02) & 0.92 (0.02) & 0.92 (0.03) & 0.99 (0.01) & 0.99 (0.01) & 0.99 (0.01)\\
 & 45 & 0.90 (0.06) & 0.90 (0.06) & 0.90 (0.06) & 0.99 (0.01) & 0.99 (0.01) & 0.99 (0.01)\\
 & 50 & 0.93 (0.04) & 0.93 (0.04) & -- & 1.00 (0.01) & 1.00 (0.01) & --\\
\hline\addlinespace
1 & 20 & 0.89 (0.07) & 0.89 (0.07) & 0.89 (0.07) & 1.00 (0.00) & 0.99 (0.02) & 0.99 (0.02)\\
 & 25 & 0.94 (0.04) & 0.93 (0.03) & 0.93 (0.03) & 0.99 (0.02) & 0.99 (0.02) & 0.99 (0.02)\\
 & 30 & 0.94 (0.08) & 0.94 (0.08) & 0.94 (0.08) & 1.00 (0.00) & 1.00 (0.00) & 1.00 (0.00)\\
 & 35 & 0.91 (0.06) & 0.91 (0.06) & 0.91 (0.06) & 1.00 (0.00) & 1.00 (0.00) & 1.00 (0.00)\\
 & 40 & 0.93 (0.04) & 0.93 (0.04) & 0.93 (0.04) & 0.99 (0.01) & 0.99 (0.01) & 0.99 (0.01)\\
 & 45 & 0.90 (0.03) & 0.90 (0.03) & 0.90 (0.03) & 1.00 (0.01) & 1.00 (0.01) & 1.00 (0.01)\\
 & 50 & 0.94 (0.02) & 0.94 (0.03) & -- & 1.00 (0.00) & 1.00 (0.00) & --\\
\hline\addlinespace
5 & 20 & 0.92 (0.07) & 0.92 (0.07) & 0.92 (0.07) & 0.99 (0.02) & 0.99 (0.02) & 0.99 (0.02)\\
 & 25 & 0.92 (0.05) & 0.92 (0.05) & 0.92 (0.05) & 1.00 (0.00) & 1.00 (0.00) & 1.00 (0.00)\\
 & 30 & 0.87 (0.04) & 0.87 (0.03) & 0.87 (0.03) & 1.00 (0.00) & 1.00 (0.00) & 1.00 (0.00)\\
 & 35 & 0.94 (0.04) & 0.93 (0.03) & 0.93 (0.03) & 1.00 (0.00) & 1.00 (0.00) & 1.00 (0.00)\\
 & 40 & 0.94 (0.04) & 0.94 (0.04) & 0.94 (0.04) & 0.99 (0.01) & 0.99 (0.01) & 0.99 (0.01)\\
 & 45 & 0.92 (0.06) & 0.92 (0.06) & 0.92 (0.06) & 1.00 (0.01) & 1.00 (0.01) & 1.00 (0.01)\\
 & 50 & 0.94 (0.02) & 0.93 (0.03) & -- & 0.99 (0.01) & 0.99 (0.01) & --\\
\bottomrule
\end{tabular}}
\end{table}
 
\begin{table}[htb!]

\caption{\label{tab:binomial_fss_time}Simulation without multicollinearity: Comparison of different solvers and complete enumeration for logistic regression; average execution times in seconds.}
\centering
\resizebox{\ifdim\width>\linewidth\linewidth\else\width\fi}{!}{
\begin{tabular}[t]{>{}r|rr>{}r|rrr}
\toprule
\multicolumn{7}{c}{Logistic model training time} \\
\cmidrule(l{3pt}r{3pt}){1-7}
\multicolumn{1}{c}{ } & \multicolumn{3}{c}{AIC} & \multicolumn{3}{c}{BIC} \\
\cmidrule(l{3pt}r{3pt}){2-4} \cmidrule(l{3pt}r{3pt}){5-7}
\multicolumn{1}{c}{ } & \multicolumn{1}{c}{Proposed Method} & \multicolumn{1}{c}{Proposed Method} & \multicolumn{1}{c}{ } & \multicolumn{1}{c}{Proposed Method} & \multicolumn{1}{c}{Proposed Method} & \multicolumn{1}{c}{ } \\
\multicolumn{1}{c}{p} & \multicolumn{1}{c}{(ECOS)} & \multicolumn{1}{c}{(MOSEK)} & \multicolumn{1}{c}{bestglm} & \multicolumn{1}{c}{(ECOS)} & \multicolumn{1}{c}{(MOSEK)} & \multicolumn{1}{c}{bestglm} \\
\cmidrule(l{3pt}r{3pt}){1-1} \cmidrule(l{3pt}r{3pt}){2-2} \cmidrule(l{3pt}r{3pt}){3-3} \cmidrule(l{3pt}r{3pt}){4-4} \cmidrule(l{3pt}r{3pt}){5-5} \cmidrule(l{3pt}r{3pt}){6-6} \cmidrule(l{3pt}r{3pt}){7-7}
5 & 0.896 & 2.329 & 0.081 & 0.878 & 2.188 & 0.081\\
10 & 1.971 & 4.244 & 2.522 & 2.035 & 4.188 & 2.510\\
15 & 4.553 & 7.220 & 94.975 & 3.076 & 5.809 & 95.105\\
20 & 12.635 & 19.999 & -- & 5.051 & 11.751 & --\\
25 & 21.486 & 29.610 & -- & 7.755 & 13.668 & --\\
30 & 27.868 & 38.201 & -- & 11.001 & 14.549 & --\\
35 & 77.166 & 105.110 & -- & 21.279 & 25.351 & --\\
40 & 47.305 & 91.766 & -- & 33.676 & 28.208 & --\\
\bottomrule
\end{tabular}}
\end{table}
 \begin{table}[htb!]

\caption{\label{tab:binomial_fss_sel}Simulation without multicollinearity: Comparison of different solvers and complete enumeration for logistic regression; average (sd) selection accuracy.}
\centering
\resizebox{\ifdim\width>\linewidth\linewidth\else\width\fi}{!}{
\begin{tabular}[t]{>{}r|rr>{}r|rrr}
\toprule
\multicolumn{7}{c}{Logistic model support recovery accuracy} \\
\cmidrule(l{3pt}r{3pt}){1-7}
\multicolumn{1}{c}{ } & \multicolumn{3}{c}{AIC} & \multicolumn{3}{c}{BIC} \\
\cmidrule(l{3pt}r{3pt}){2-4} \cmidrule(l{3pt}r{3pt}){5-7}
\multicolumn{1}{c}{ } & \multicolumn{1}{c}{Proposed Method} & \multicolumn{1}{c}{Proposed Method} & \multicolumn{1}{c}{ } & \multicolumn{1}{c}{Proposed Method} & \multicolumn{1}{c}{Proposed Method} & \multicolumn{1}{c}{ } \\
\multicolumn{1}{c}{p} & \multicolumn{1}{c}{(ECOS)} & \multicolumn{1}{c}{(MOSEK)} & \multicolumn{1}{c}{bestglm} & \multicolumn{1}{c}{(ECOS)} & \multicolumn{1}{c}{(MOSEK)} & \multicolumn{1}{c}{bestglm} \\
\cmidrule(l{3pt}r{3pt}){1-1} \cmidrule(l{3pt}r{3pt}){2-2} \cmidrule(l{3pt}r{3pt}){3-3} \cmidrule(l{3pt}r{3pt}){4-4} \cmidrule(l{3pt}r{3pt}){5-5} \cmidrule(l{3pt}r{3pt}){6-6} \cmidrule(l{3pt}r{3pt}){7-7}
5 & 1.00 (0.00) & 1.00 (0.00) & 1.00 (0.00) & 1.00 (0.00) & 1.00 (0.00) & 1.00 (0.00)\\
10 & 0.94 (0.06) & 0.94 (0.06) & 0.94 (0.06) & 0.98 (0.04) & 0.98 (0.04) & 0.98 (0.04)\\
15 & 0.95 (0.03) & 0.95 (0.03) & 0.95 (0.03) & 1.00 (0.00) & 1.00 (0.00) & 1.00 (0.00)\\
20 & 0.93 (0.03) & 0.93 (0.03) & -- & 1.00 (0.00) & 1.00 (0.00) & --\\
25 & 0.92 (0.07) & 0.92 (0.07) & -- & 0.99 (0.02) & 0.99 (0.02) & --\\
30 & 0.95 (0.04) & 0.95 (0.04) & -- & 1.00 (0.00) & 1.00 (0.00) & --\\
35 & 0.95 (0.02) & 0.95 (0.02) & -- & 1.00 (0.00) & 1.00 (0.00) & --\\
40 & 0.95 (0.02) & 0.95 (0.02) & -- & 0.99 (0.01) & 0.99 (0.01) & --\\
\bottomrule
\end{tabular}}
\end{table}
 
\begin{table}[htb!]

\caption{\label{tab:poisson_fss_time}Simulation without multicollinearity: Comparison of different solvers and complete enumeration for Poisson regression; average execution times in seconds.}
\centering
\resizebox{\ifdim\width>\linewidth\linewidth\else\width\fi}{!}{
\begin{tabular}[t]{>{}r|rr>{}r|rrr}
\toprule
\multicolumn{7}{c}{Poisson model training time} \\
\cmidrule(l{3pt}r{3pt}){1-7}
\multicolumn{1}{c}{ } & \multicolumn{3}{c}{AIC} & \multicolumn{3}{c}{BIC} \\
\cmidrule(l{3pt}r{3pt}){2-4} \cmidrule(l{3pt}r{3pt}){5-7}
\multicolumn{1}{c}{ } & \multicolumn{1}{c}{Proposed Method} & \multicolumn{1}{c}{Proposed Method} & \multicolumn{1}{c}{ } & \multicolumn{1}{c}{Proposed Method} & \multicolumn{1}{c}{Proposed Method} & \multicolumn{1}{c}{ } \\
\multicolumn{1}{c}{p} & \multicolumn{1}{c}{(ECOS)} & \multicolumn{1}{c}{(MOSEK)} & \multicolumn{1}{c}{bestglm} & \multicolumn{1}{c}{(ECOS)} & \multicolumn{1}{c}{(MOSEK)} & \multicolumn{1}{c}{bestglm} \\
\cmidrule(l{3pt}r{3pt}){1-1} \cmidrule(l{3pt}r{3pt}){2-2} \cmidrule(l{3pt}r{3pt}){3-3} \cmidrule(l{3pt}r{3pt}){4-4} \cmidrule(l{3pt}r{3pt}){5-5} \cmidrule(l{3pt}r{3pt}){6-6} \cmidrule(l{3pt}r{3pt}){7-7}
5 & 1.962 & 1.290 & 0.141 & 1.868 & 1.229 & 0.141\\
10 & 6.847 & 3.052 & 3.610 & 5.667 & 3.061 & 3.619\\
15 & 13.111 & 5.903 & 141.839 & 12.000 & 5.562 & 141.700\\
20 & 35.552 & 11.577 & -- & 34.650 & 9.628 & --\\
25 & 238.299 & 18.425 & -- & 220.857 & 12.383 & --\\
30 & 896.185 & 49.401 & -- & 952.785 & 20.371 & --\\
35 & 2408.308 & 51.174 & -- & 2346.494 & 33.675 & --\\
40 & 4030.605 & 215.312 & -- & 3354.529 & 35.426 & --\\
\bottomrule
\end{tabular}}
\end{table}
 \begin{table}[htb!]

\caption{\label{tab:poisson_fss_sel}Simulation without multicollinearity: Comparison of different solvers and complete enumeration for Poisson regression; average (sd) selection accuracy.}
\centering
\resizebox{\ifdim\width>\linewidth\linewidth\else\width\fi}{!}{
\begin{tabular}[t]{>{}r|rr>{}r|rrr}
\toprule
\multicolumn{7}{c}{Poisson model support recovery accuracy} \\
\cmidrule(l{3pt}r{3pt}){1-7}
\multicolumn{1}{c}{ } & \multicolumn{3}{c}{AIC} & \multicolumn{3}{c}{BIC} \\
\cmidrule(l{3pt}r{3pt}){2-4} \cmidrule(l{3pt}r{3pt}){5-7}
\multicolumn{1}{c}{ } & \multicolumn{1}{c}{Proposed Method} & \multicolumn{1}{c}{Proposed Method} & \multicolumn{1}{c}{ } & \multicolumn{1}{c}{Proposed Method} & \multicolumn{1}{c}{Proposed Method} & \multicolumn{1}{c}{ } \\
\multicolumn{1}{c}{p} & \multicolumn{1}{c}{(ECOS)} & \multicolumn{1}{c}{(MOSEK)} & \multicolumn{1}{c}{bestglm} & \multicolumn{1}{c}{(ECOS)} & \multicolumn{1}{c}{(MOSEK)} & \multicolumn{1}{c}{bestglm} \\
\cmidrule(l{3pt}r{3pt}){1-1} \cmidrule(l{3pt}r{3pt}){2-2} \cmidrule(l{3pt}r{3pt}){3-3} \cmidrule(l{3pt}r{3pt}){4-4} \cmidrule(l{3pt}r{3pt}){5-5} \cmidrule(l{3pt}r{3pt}){6-6} \cmidrule(l{3pt}r{3pt}){7-7}
5 & 0.96 (0.09) & 0.96 (0.09) & 0.96 (0.09) & 0.92 (0.18) & 1.00 (0.00) & 1.00 (0.00)\\
10 & 0.92 (0.04) & 0.92 (0.04) & 0.92 (0.04) & 1.00 (0.00) & 1.00 (0.00) & 1.00 (0.00)\\
15 & 0.87 (0.09) & 0.87 (0.09) & 0.87 (0.09) & 1.00 (0.00) & 1.00 (0.00) & 1.00 (0.00)\\
20 & 0.97 (0.07) & 0.94 (0.06) & -- & 1.00 (0.00) & 1.00 (0.00) & --\\
25 & 0.90 (0.21) & 0.95 (0.04) & -- & 0.90 (0.21) & 1.00 (0.00) & --\\
30 & 0.87 (0.20) & 0.94 (0.06) & -- & 0.90 (0.19) & 1.00 (0.00) & --\\
35 & 0.81 (0.22) & 0.93 (0.04) & -- & 0.82 (0.23) & 0.99 (0.02) & --\\
40 & 0.67 (0.26) & 0.94 (0.03) & -- & 0.78 (0.28) & 0.99 (0.01) & --\\
\bottomrule
\end{tabular}}
\end{table}
 
\begin{table}[htb!]

\caption{
    Simulation with multicollinearity:
    Comparison of the impact of different pairwise correlation constraints,
    for different levels of correlation, on the MSE for the logistic regression model.
    \label{tab:sim2_binomial}
}
\centering
\resizebox{\ifdim\width>\linewidth\linewidth\else\width\fi}{!}{\begin{tabular}[t]{lr>{\raggedright\arraybackslash}p{2.5cm}>{\raggedright\arraybackslash}p{2.5cm}>{\raggedright\arraybackslash}p{2.5cm}>{\raggedright\arraybackslash}p{2.5cm}>{\raggedright\arraybackslash}p{2.5cm}}
\toprule
\multicolumn{2}{c}{ } & \multicolumn{5}{c}{Logistic model mean-squared error} \\
\cmidrule(l{3pt}r{3pt}){3-7}
IC & alpha & no constraint & at most one constraint & sign coherence 0.5 constraint & sign coherence 0.7 constraint & combined constraint\\
\midrule
AIC & 0.70 & 1.41e-01 & 1.41e-01 & 1.41e-01 & 1.41e-01 & \textbf{1.40e-01}\\
AIC & 0.80 & 1.96e-01 & 2.20e-01 & 1.96e-01 & 1.96e-01 & \textbf{1.55e-01}\\
AIC & 0.90 & 3.13e-01 & 4.94e-01 & 3.07e-01 & 3.07e-01 & \textbf{3.07e-02}\\
AIC & 0.95 & 5.18e-01 & 5.66e-01 & 4.90e-01 & 4.90e-01 & \textbf{2.95e-02}\\
AIC & 0.99 & 1.21e+00 & 6.20e-01 & 6.20e-01 & 6.20e-01 & \textbf{2.89e-02}\\
BIC & 0.70 & 2.10e-01 & 2.10e-01 & 2.10e-01 & 2.10e-01 & \textbf{2.10e-01}\\
BIC & 0.80 & 2.75e-01 & 2.89e-01 & 2.75e-01 & 2.75e-01 & \textbf{2.22e-01}\\
BIC & 0.90 & 4.50e-01 & 4.94e-01 & 4.50e-01 & 4.50e-01 & \textbf{3.10e-02}\\
BIC & 0.95 & 5.67e-01 & 5.66e-01 & 5.64e-01 & 5.64e-01 & \textbf{2.96e-02}\\
BIC & 0.99 & 7.68e-01 & 6.20e-01 & 6.20e-01 & 6.20e-01 & \textbf{2.89e-02}\\
\bottomrule
\end{tabular}}
\end{table}
 \begin{table}[htb!]

\caption{
    Simulation with multicollinearity:
    Comparison of the impact of different pairwise correlation constraints,
    for different levels of correlation, on the MSE for the Poisson regression model.
    \label{tab:sim2_poisson}
}
\centering
\resizebox{\ifdim\width>\linewidth\linewidth\else\width\fi}{!}{
\begin{tabular}[t]{lr>{\raggedright\arraybackslash}p{2.5cm}>{\raggedright\arraybackslash}p{2.5cm}>{\raggedright\arraybackslash}p{2.5cm}>{\raggedright\arraybackslash}p{2.5cm}>{\raggedright\arraybackslash}p{2.5cm}}
\toprule
\multicolumn{2}{c}{ } & \multicolumn{5}{c}{Poisson model mean-squared error} \\
\cmidrule(l{3pt}r{3pt}){3-7}
IC & alpha & no constraint & at most one constraint & sign coherence 0.5 constraint & sign coherence 0.7 constraint & combined constraint\\
\midrule
AIC & 0.70 & 8.98e-03 & 9.50e-03 & 8.98e-03 & 8.98e-03 & \textbf{8.97e-03}\\
AIC & 0.80 & 9.54e-03 & 5.81e-02 & 9.54e-03 & 9.54e-03 & \textbf{8.51e-03}\\
AIC & 0.90 & 1.36e-02 & 3.93e-01 & 1.36e-02 & 1.36e-02 & \textbf{4.13e-03}\\
AIC & 0.95 & 2.71e-02 & 4.42e-01 & 2.71e-02 & 2.71e-02 & \textbf{3.11e-03}\\
AIC & 0.99 & 1.58e-01 & 4.95e-01 & 1.57e-01 & 1.57e-01 & \textbf{5.77e-03}\\
BIC & 0.70 & 8.98e-03 & 9.50e-03 & 8.98e-03 & 8.98e-03 & \textbf{8.97e-03}\\
BIC & 0.80 & 9.66e-03 & 5.83e-02 & 9.66e-03 & 9.66e-03 & \textbf{8.56e-03}\\
BIC & 0.90 & 1.59e-02 & 3.93e-01 & 1.59e-02 & 1.59e-02 & \textbf{4.70e-03}\\
BIC & 0.95 & 4.54e-02 & 4.42e-01 & 4.54e-02 & 4.54e-02 & \textbf{3.11e-03}\\
BIC & 0.99 & 2.42e-01 & 4.95e-01 & 2.41e-01 & 2.41e-01 & \textbf{4.98e-03}\\
\bottomrule
\end{tabular}}
\end{table}
 \begin{table}[htb!]

\caption{
    Simulation with multicollinearity:
    Comparison of the impact of different pairwise correlation constraints,
    for different levels of correlation, on the MSE for the linear regression model.
    \label{tab:sim2_gaussian}
}
\centering
\resizebox{\ifdim\width>\linewidth\linewidth\else\width\fi}{!}{
\begin{tabular}[t]{rr>{\raggedright\arraybackslash}p{2.5cm}>{\raggedright\arraybackslash}p{2.5cm}>{\raggedright\arraybackslash}p{2.5cm}>{\raggedright\arraybackslash}p{2.5cm}>{\raggedright\arraybackslash}p{2.5cm}}
\toprule
\multicolumn{1}{c}{ } & \multicolumn{1}{c}{ } & \multicolumn{5}{c}{Linear model mean-squared error} \\
\cmidrule(l{3pt}r{3pt}){3-7}
alpha & sd & no constraint & at most one constraint & sign coherence 0.5 constraint & sign coherence 0.7 constraint & combined constraint\\
\midrule
0.70 & 0.01 & \textbf{1.46e-06} & 5.04e-04 & \textbf{1.46e-06} & \textbf{1.46e-06} & 1.57e-06\\
0.70 & 0.10 & 1.46e-04 & 6.45e-04 & 1.46e-04 & 1.46e-04 & \textbf{1.45e-04}\\
0.70 & 0.20 & 5.84e-04 & 1.08e-03 & 5.84e-04 & 5.84e-04 & \textbf{5.81e-04}\\
0.70 & 0.30 & 1.31e-03 & 1.81e-03 & 1.31e-03 & 1.31e-03 & \textbf{1.31e-03}\\
0.70 & 0.40 & 2.33e-03 & 2.83e-03 & 2.33e-03 & 2.33e-03 & \textbf{2.32e-03}\\
0.70 & 0.50 & 3.65e-03 & 4.14e-03 & 3.65e-03 & 3.65e-03 & \textbf{3.63e-03}\\
0.70 & 1.00 & 1.48e-02 & 1.52e-02 & 1.46e-02 & 1.48e-02 & \textbf{1.46e-02}\\
0.80 & 0.01 & \textbf{1.90e-06} & 4.71e-02 & \textbf{1.90e-06} & \textbf{1.90e-06} & 1.18e-05\\
0.80 & 0.10 & 1.90e-04 & 4.72e-02 & 1.90e-04 & 1.90e-04 & \textbf{1.57e-04}\\
0.80 & 0.20 & 7.59e-04 & 4.79e-02 & 7.59e-04 & 7.59e-04 & \textbf{5.97e-04}\\
0.80 & 0.30 & 1.71e-03 & 4.88e-02 & 1.71e-03 & 1.71e-03 & \textbf{1.33e-03}\\
0.80 & 0.40 & 3.04e-03 & 5.02e-02 & 3.04e-03 & 3.04e-03 & \textbf{2.36e-03}\\
0.80 & 0.50 & 4.75e-03 & 5.18e-02 & 4.75e-03 & 4.75e-03 & \textbf{3.67e-03}\\
0.80 & 1.00 & 1.98e-02 & 6.63e-02 & 1.90e-02 & 1.94e-02 & \textbf{1.47e-02}\\
0.90 & 0.01 & \textbf{3.27e-06} & 3.86e-01 & \textbf{3.27e-06} & \textbf{3.27e-06} & 1.69e-05\\
0.90 & 0.10 & 3.27e-04 & 3.87e-01 & 3.27e-04 & 3.27e-04 & \textbf{5.07e-05}\\
0.90 & 0.20 & 1.31e-03 & 3.88e-01 & 1.31e-03 & 1.31e-03 & \textbf{1.53e-04}\\
0.90 & 0.30 & 2.94e-03 & 3.90e-01 & 2.94e-03 & 2.94e-03 & \textbf{3.24e-04}\\
0.90 & 0.40 & 5.23e-03 & 3.92e-01 & 5.23e-03 & 5.23e-03 & \textbf{5.63e-04}\\
0.90 & 0.50 & 8.17e-03 & 3.95e-01 & 8.17e-03 & 8.17e-03 & \textbf{8.70e-04}\\
0.90 & 1.00 & 4.52e-02 & 4.09e-01 & 3.25e-02 & 3.25e-02 & \textbf{3.43e-03}\\
0.95 & 0.01 & 6.04e-06 & 4.39e-01 & 6.04e-06 & 6.04e-06 & \textbf{3.86e-06}\\
0.95 & 0.10 & 6.04e-04 & 4.40e-01 & 6.04e-04 & 6.04e-04 & \textbf{3.70e-05}\\
0.95 & 0.20 & 2.42e-03 & 4.41e-01 & 2.42e-03 & 2.42e-03 & \textbf{1.38e-04}\\
0.95 & 0.30 & 5.44e-03 & 4.43e-01 & 5.44e-03 & 5.44e-03 & \textbf{3.05e-04}\\
0.95 & 0.40 & 9.67e-03 & 4.44e-01 & 9.67e-03 & 9.67e-03 & \textbf{5.40e-04}\\
0.95 & 0.50 & 1.55e-02 & 4.46e-01 & 1.51e-02 & 1.51e-02 & \textbf{8.42e-04}\\
0.95 & 1.00 & 1.05e-01 & 4.57e-01 & 5.85e-02 & 5.85e-02 & \textbf{3.36e-03}\\
0.99 & 0.01 & 2.83e-05 & 4.87e-01 & 2.83e-05 & 2.83e-05 & \textbf{4.56e-07}\\
0.99 & 0.10 & 2.83e-03 & 4.88e-01 & 2.83e-03 & 2.83e-03 & \textbf{3.32e-05}\\
0.99 & 0.20 & 1.14e-02 & 4.88e-01 & 1.13e-02 & 1.13e-02 & \textbf{1.32e-04}\\
0.99 & 0.30 & 3.25e-02 & 4.89e-01 & 2.54e-02 & 2.54e-02 & \textbf{2.98e-04}\\
0.99 & 0.40 & 7.63e-02 & 4.91e-01 & 4.43e-02 & 4.43e-02 & \textbf{5.29e-04}\\
0.99 & 0.50 & 1.26e-01 & 4.92e-01 & 6.75e-02 & 6.75e-02 & \textbf{8.27e-04}\\
0.99 & 1.00 & 4.04e-01 & 4.99e-01 & 2.10e-01 & 2.10e-01 & \textbf{3.31e-03}\\
\bottomrule
\end{tabular}}
\end{table}

\clearpage

\bibliographystyle{elsarticle-num-names} 
\bibliography{aglm}

\end{document}